 \documentclass[pmlr,twocolumn,10pt]{jmlr} % W&CP article

\usepackage{booktabs}
\usepackage[load-configurations=version-1]{siunitx} % newer version 
\theorembodyfont{\upshape}
\theoremheaderfont{\scshape}
\theorempostheader{:}
\theoremsep{\newline}

\jmlrvolume{LEAVE UNSET}
\jmlryear{2021}
\jmlrsubmitted{LEAVE UNSET}
\jmlrpublished{LEAVE UNSET}
\jmlrworkshop{Machine Learning for Health (ML4H) 2021} % W&CP title

\title[CARSA]{Brain
dynamics via Cumulative Auto-Regressive Self-Attention}

  \author{%
   \Name{Usman Mahmood} \Email{umahmood1@student.gsu.edu}\\
   \Name{Zening Fu} \Email{zfu@gsu.edu}\\
   \Name{Vince Calhoun} \Email{vcalhoun.gsu.edu}\\
   \Name{Sergey Plis} \Email{s.m.plis@gmail.com}\\
   \addr Tri-institutional Center for Translational Research in Neuroimaging and Data Science (TReNDS), Georgia State University, Georgia Tech, Emory
  }

\begin{document}

\maketitle

\begin{abstract}
 Multivariate dynamical processes can often be intuitively described by a weighted connectivity graph between components representing each individual time-series. Even a simple representation of this graph as a Pearson correlation matrix may be informative and predictive as demonstrated in the brain imaging literature.
  However, there is a consensus expectation that powerful graph neural networks (GNNs) should perform better in similar settings.
  In this work we present a model that is considerably shallow than deep GNNs, yet outperforms them in predictive accuracy in a brain imaging application.
  Our model learns the autoregressive structure of individual time series and estimates directed connectivity graphs between the learned representations via a self-attention mechanism in an end-to-end fashion. The supervised training of the model as a classifier between patients and controls results in a model that generates directed connectivity graphs and highlights the components of the time-series that are predictive for each subject. We demonstrate our results on a functional neuroimaging dataset classifying schizophrenia patients and controls.
  \end{abstract}
\begin{keywords}
Structure Learning, functional MRI, Schizophrenia, Classification
\end{keywords}

\section{Introduction}
\label{sec:intro}

Study and predictive diagnostics of mental disorders have been a growing area of research~\cite{Lynall9477, khosla2019machine,frontiers2014, 2021}.
The research focuses on building classification models for such disorders in the hope to learn disease specific regions and their interaction.
Abnormal function of specific brain regions is often a characteristic of specific mental disorder of the subject \cite{article, https://doi.org/10.1002/hbm.25205, article2, article3}. Often these two research problems---classification and brain region interactions (FNC)---are handled separately.
Approaches that create a single method to handle these two problems simultaneously remain vastly unexplored.

In this paper, we use resting state functional magnetic resonance imaging (rs-fMRI)---spatio-temporal brain data---and present an approach that simultaneously classifies schizophrenia subjects from healthy controls, and learns the effective connectivity structure (graph) among different components/regions.

We present an approach to learn brain regions' interaction as a graph learning method. In this graph, different fMRI regions/components represent nodes and their interaction/connectivity serve as edge weights. As, the underlying ground truth graph structure is missing, our method could be considered an un-supervised graph learning approach. However, as the training is supervised, although to a different task, it is more precisely classified as a self-supervised method. Our proposed approach can also be considered as graphical representation learning. The input data consist of components created using independent component analysis (ICA) on rs-fMRI data (see Section~\ref{Dataset}).

Existing unsupervised and self-supervised graph learning methods~\cite{kipf2018neural, shang2021discrete} often rely on predicting future values in time of the components/objects in the space they are defined using the learned graph structure. Learning is performed by maximizing similarity between prediction and the ground truth time courses. The problem with these approaches, especially for fMRI, is that predicting future very far in time is significantly more difficult for real than for simulated data because of the presence of noise, and many features which are responsible of future values, are often missing from input data. Also, the variance in input values is not that large. As shown in the experiments of~\cite{kipf2018neural}, one gets a very small error by just keeping future values equal to current value. The error is even smaller for fMRI data as the variance is of much smaller magnitude than the simulated data used in~\cite{kipf2018neural}. Lastly, the effect of the learned graph on classification remains unknown by just decreasing prediction loss, and because of un-supervised learning the efficacy of the learned graph in terms of classification remains unclear as the ground truth graph is not available.

We turn the problem into graph classification, and with high classification accuracy we can rely on the learned graph structure. Since classifying mental disorders is a significantly more difficult task than predicting few time points in future, the learned graph structure can be trusted with higher confidence. Finally, we use a selection method to select a sparse set of important nodes (referred as components/regions in the paper) for classification. Recently, there has been a surge of research in GNNs and many models have been proposed \cite{zhang2018link,PARISOT2018117,kazi2018selfattention,kipf2017semisupervised,9336270,NEURIPS2018_e77dbaf6,zhang2018end}. These models are used for different tasks and can be easily modified for graph classification. GNN models are often deep and have multiple GNN layers/steps, to gather information from distant neighbors. Most of the GNNs models used for graph classification task (GCT) are supervised and assume that the underline true graph structure of data is available, which is highly unlikely in many cases. Such models can be used by assuming a complete graph, but can greatly impact classification performance.

In this paper, we show that to simultaneously learn and classify a graph structure, a traditional deep GNN model is not required. We present a relatively shallow model (in terms of learned parameters) which demonstrates remarkable performance. Our method relies on cumulative auto-regressive representations obtained from individual time series using a recurrent neural network, which are subsequently recombined via a self-attention mechanisms. We call our model CARSA for cumulative auto-regressive self-attention. An application to an fMRI dataset demonstrates its efficacy.

\section{CARSA architecture}
\label{Arch}

We present CARSA as an end-to-end learning architecture based on classification to learn functional connectivity between components. Our model first learns the effective connectivity between time courses of an individual component's sequence. Learning the connectivity helps in creating a single embedding of the entire sequence which incorporates the effects a sequence through time. We use bidirectional long-short term memory (biLSTM) \cite{650093} to learn these relations between time courses occurring in succession. With the learned embeddings of the sequences we apply the same concept which we used across time courses to sequence embeddings. 

\begin{figure}[t]
\centering
\includegraphics[scale=0.9, width=1\columnwidth]{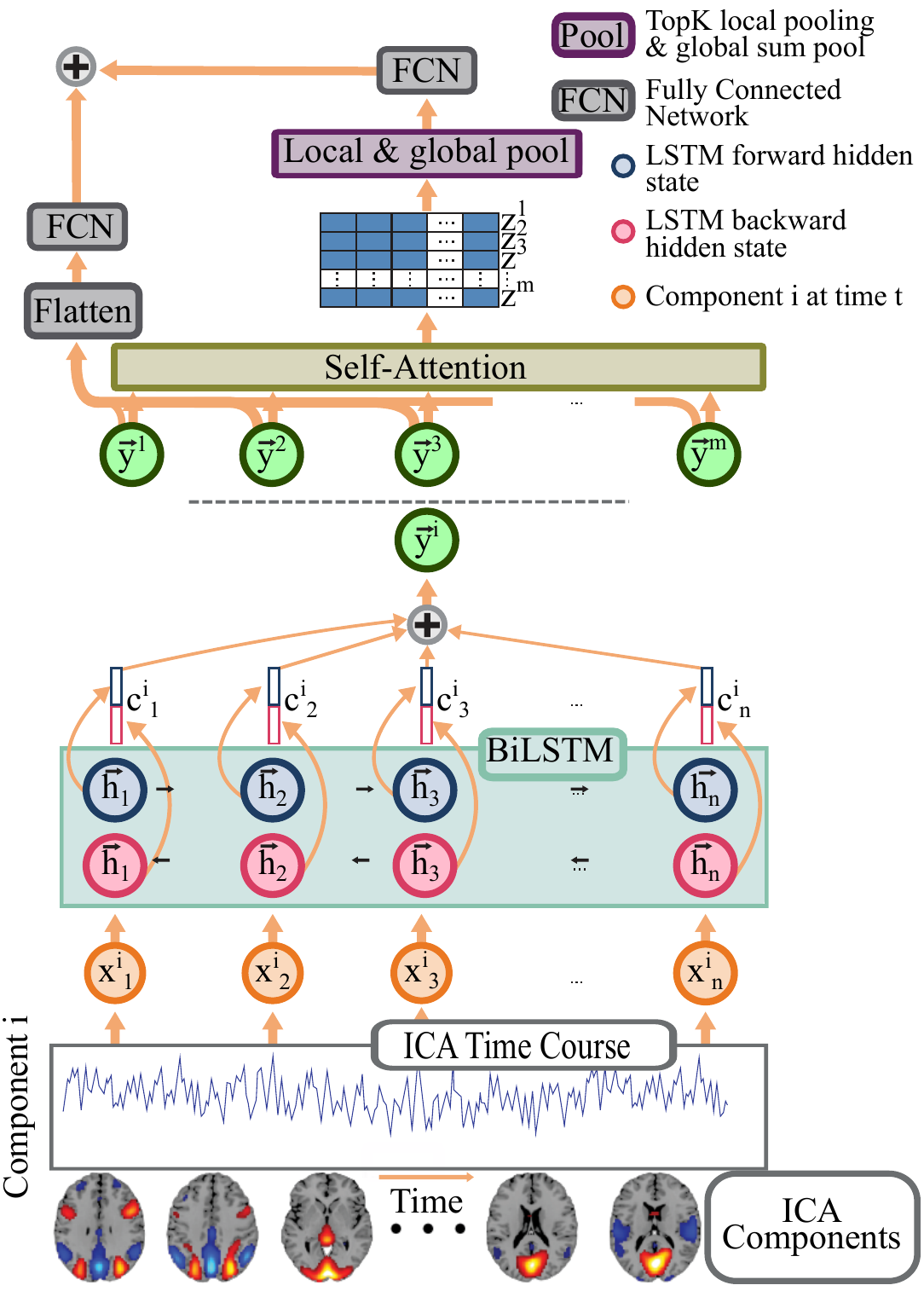}
\caption{CARSA architecture. The architecture has three distinct parts. 1) LSTM, 2) Self-Attention and 3) Pooling. }
\label{fig:LSTMArch}
\end{figure}

Unlike time courses of a sequence, different sequences are not successive, rather a sequence can have relation with any single or multiple other sequences. The connectivity between sequences can be represented as functional connectivity or graph edge weights and we capture this connectivity via self-attention based model presented in \cite{10.5555/3295222.3295349}. Finally, to acquire subject specific discriminating sequences we use a learnable pooling method \cite{gao2019graph, knyazev2019understanding}. In our experiments we use encoded rsfMRI ICA time courses as our sequences, with each subject having multiple components. Refer to Figure \ref{fig:LSTMArch} for complete architecture of our model. We explain the architecture details including hyper-parameters and importance of each important part in further sections. Our architecture is relatively a shallow network which shows high performance which we consider as an advantage over very deep neural networks. 

\subsection{LSTM for Single Components}

We use a single layer biLSTM with hidden dimension of size $64$. The LSTM receives $x^i_t$ at each time step. $x^i_t$ represents the ICA value of component $i$ at time step $t$. $x^i_t$ is dependant on many factors, one of them being the values of $x^i_{1 ... t-1}$. These relationships are very hard to capture and unlike many other time series, can't be computed using a fixed method or formula. Also, it is unknown how farther in time a component's effect remains in the time series. Learning the effective connectivity helps in identifying different time series which eventually leads to better classification. LSTMs with the help of memory and forget gates learns all these unknown factors and relationships. We learn these factors by using an end-to-end learning by classifying subjects. 
The two hidden vectors of the LSTM at step $s$ representing forward and backward passes are concatenated to create a single vector $c^i_{t}$ for component $i$ at time $t$. With $f_\theta$ representing LSTM with parameters $\theta$. This vector can be presented as:

\begin{align*}
  c^i_t = f_\theta (x^i_t \ | \ f_\theta(x^i_j) \ ; \ j = 1 \to   t-1)  
\end{align*}

Finally, to obtain a single vector for the complete sequence/time series for component $i$ we sum all the vectors $c^i$'$s$.
\\

\centerline {$y^i = \sum_{t=1}^n  c^i_t$}

%  LR: Logistic Regression, MLP: Multilayer perceptron. SVM: Support Vector Machine. The horizontal lines on top show statistical significance in difference between two methods.

% \begin{figure*}[t]
% \centering
% \includegraphics{images/AUC_FBIRN_ICA_Comparision.pdf}
% \caption{We show the mean and median of all the trials across the folds of cross validation. Our model performs better than many machine learning and deep learning methods and only $\sim 0.045$ less than the best performing model. LR: Logistic Regression, MLP: Multilayer perceptron. SVM: Support Vector Machine. The horizontal lines on top show statistical significance in difference between two methods.}
% \label{fig:AUCcomparision}
% \end{figure*}

\subsection{Self-Attention Across Components}

To capture the effective connectivity between components we use a self-attention model inspired from \cite{10.5555/3295222.3295349}, with embedding dimension of size $64$. A self-attention module computes the weights between different components and use that to create new embeddings $z^i$ for each component $i$. As, mentioned before the components are not just affected by their previous values but also by other components. This in terms of brain signals can be seen as different brain regions affecting one other. In our data we look to find the relationship between ICA components. This is known as functional connectivity which is usually computed using Pearson product-moment correlation coefficients (PCC) method, and can be used as edge weights for graphs. We learn the functional connectivity, via an end to end learning of classification. This provides insights into how these components interact with each other in terms of downstream classification. 
With function $f_\phi$ representing the self-attention function with parameters $\phi$ The resultant vector $z^i$ representing component $i$ of a subject can be represented as:

\begin{align*}
  z^i = f_\phi (y^i \ | \ y^j \ ; \ j =\ 1 \to   m)
\end{align*}

\subsection{Pooling and FCN}

Not all components are equally discriminating for classification. The components can vary across subjects and diseases. To learn the important components we select the top 'k' components for each subject using top-k pooling \cite{gao2019graph, knyazev2019understanding}. We use 3 layers of such pooling dropping $20\%$  components at each layer. To obtain a final vector for the entire subject we use average pooling summing all the components after each top-k pooling layer. In the end we sum the three vectors to create a single vector representing a subject. 
For classification we use two fully connected layers with size $64$ and $2$.

\subsection{Dataset}
\label{Dataset}
We worked with the data from Function Biomedical Informatics Research Network (FBIRN)~\cite{keator2016function} dataset including SZ patients and HC for testing our model. Resting fMRI data from the phase III FBIRN were analyzed for this project. The dataset has $368$ total subjects.The fMRI data was preprocessed using statistical parametric mapping (SPM12, http://www.fil.ion.ucl.ac.uk/spm/) under MATLAB 2020 environment.
After the preprocessing, subjects were included in the analysis if the
subjects have head motion $\le 3^\circ$ and $\le 3$ mm, and with functional data providing near full brain successful normalization~\cite{fu2019altered}.

We selected subjects for further analysis if the subjects had head motion $\le 3^\circ$ and $\le 3$ mm, and with functional data providing near full brain successful normalization~\cite{fu2019altered}.

This result in a total of $311$ subjects with $151$ healthy controls and $160$ subjects with schizophrenia. Lastly, $100$ ICA components are acquired using the same procedure described in~\cite{fu2019altered}.

\begin{figure}[htbp]
 % Caption and label go in the first argument and the figure contents
 % go in the second argument
\floatconts
  {fig:AUCcomparision}
  {\caption{We show the mean and median of all trials across $4$ folds of cross validation. Our model out-performs ML and DL methods. }}
  {\includegraphics[width=\linewidth]{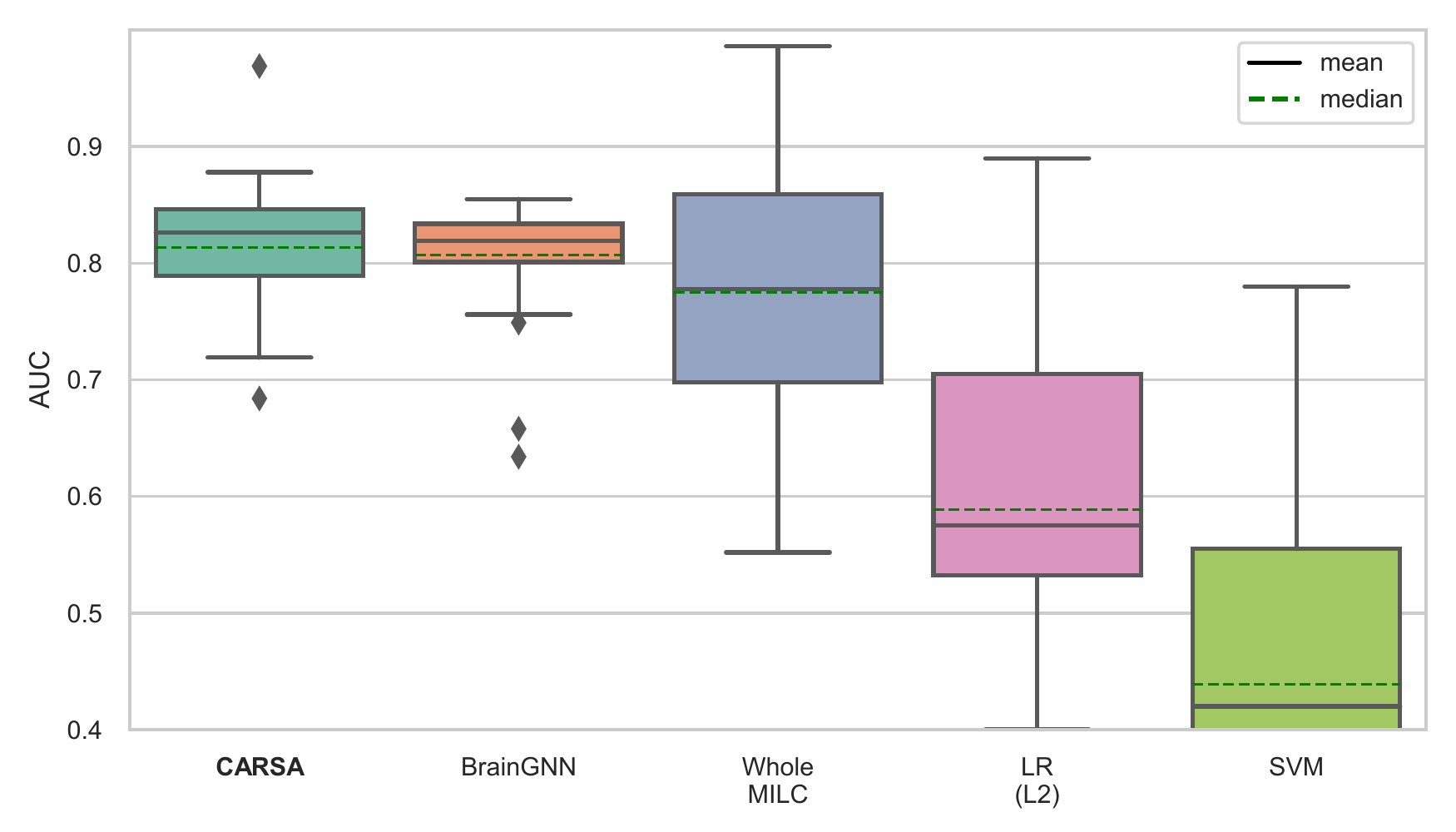}}
\end{figure}

\begin{figure}[htbp]
\floatconts
  {fig:FC}
  {\caption{\ref{fig:FNC-Important} shows the connectivity between the imporant components, grouped into 7 domains. \ref{fig:FNC-Noise} shows the FNC between noise components. Our method accurately depicts high connectivity between important components. Visual components have high connectivity with components of all other domains. SC-Subcortical, AU-Auditory, SM-Sensorimotor, VI-Visual, CC-Cognitive Control, DM-Default Mode, and CB-Cerebellum.}}
  {%
    \subfigure[FNC - Important Components]{\label{fig:FNC-Important}%
      \includegraphics[width=0.4\linewidth]{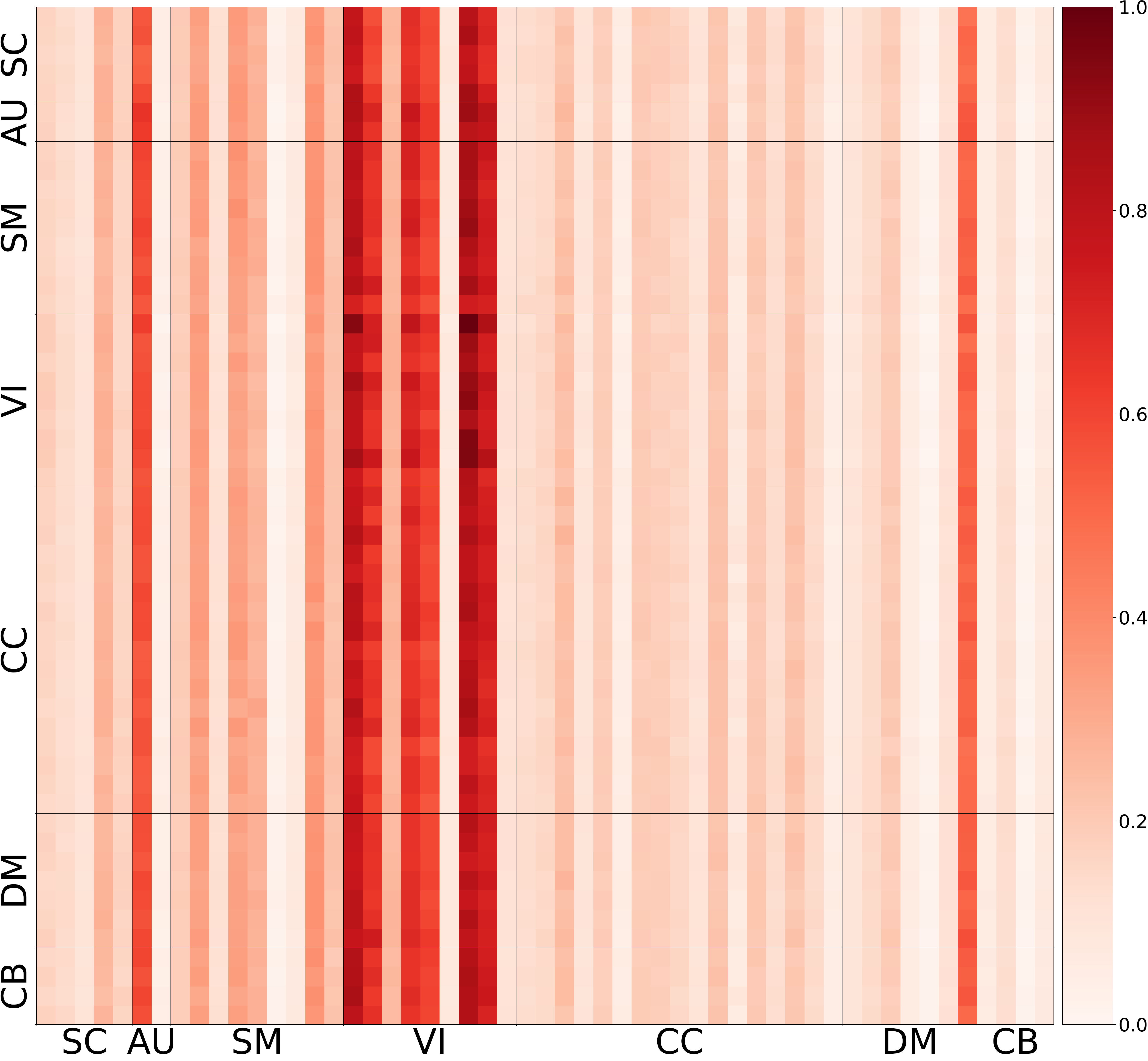}}%
    \qquad
    \subfigure[FNC - Noise Components]{\label{fig:FNC-Noise}%
      \includegraphics[width=0.4\linewidth]{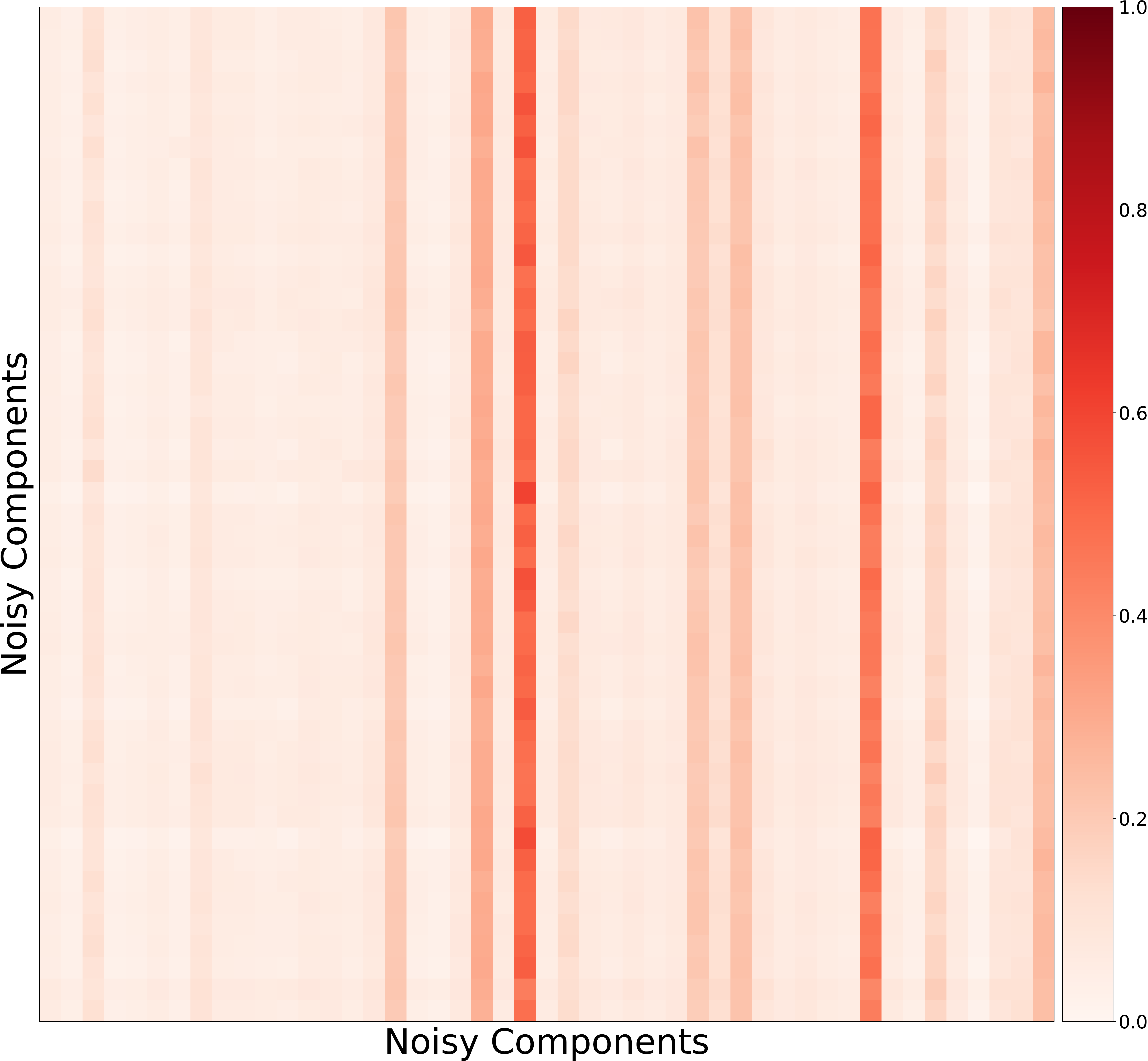}}
  }
\end{figure}

% \begin{figure}[t]
%  % Caption and label go in the first argument and the figure contents
%  % go in the second argument
% \floatconts
%   {fig:FC}
%   {\caption{Connectivity matrices between components of ICA time courses of SZ and HC subjects. The difference across disease is prominent with SZ being much sparser. We show only a few examples, as there are hundreds of subject across folds and not all subjects show such drastic difference and as mentioned before the connectivity changes across subjects as well.}}
%   {\includegraphics[width=.85\linewidth]{images/FNC-CARSA .pdf}}
% \end{figure}

% \begin{figure}[h]
% \centering
% \includegraphics[width=\linewidth]{"images/Connectivity Matrix for test Subjects"}
% \caption{Connectivity matrices between components of ICA time courses of SZ and HC subjects. The difference across disease is prominent with SZ being much sparser. We show only a few examples, as there are hundreds of subject across folds and not all subjects show such drastic difference and as mentioned before the connectivity changes across subjects as well. }
% \label{fig:FC}
% \end{figure}

\subsection{Results}
In our experiments we used 4 fold cross validation with training and validation size of $\sim 80$.
The division does not follow the common norm because of the low number of subjects, and any further reduction of validation or test data results in higher variance in the results but the average remains same. We perform $10$ randomly seed trials for each fold.

\textbf{ Classification: }The classification results in figure \ref{fig:AUCcomparision} show the efficacy of our model with performance better than machine learning deep learning methods \cite{2020,2021}. BrainGNN uses region-based data, we tried using ICA data with BrainGNN but we couldn't make it work.

\textbf{Functional connectivity: } Figure \ref{fig:FC} show the functional connectivity learned via the attention part of the model. We divide the $100$ components into important and noise category and further divide important components into $7$ domains following \cite{10.3389/fnsys.2011.00002}. These connectivity matrices can be used as the graph structure between components by either a) using the matrix directly as weights or b) creating a probabilistic graph structure using these matrices as parameters.

% \textbf{Discriminating components: } As hypothesized, in Figure \ref{fig:regions} we show different components selected by the model and compare the components between the SZ and HC groups. As show in this example taken from the first fold of test data, there is a clear difference in component selection across the underlying disorder. In more extensive experiments, this difference was seen across the folds and trials. We further saw a difference ing component selection across subjects belonging to even the same class. 

% \begin{figure}[t]
%  % Caption and label go in the first argument and the figure contents
%  % go in the second argument
% \floatconts
%   {fig:regions}
%   {\caption{Disparity in components selected for SZ and HC subjects. Figure shows the probability density of selected components. Left side has sorted components for SZ patients and right has sorted components for HC subjects.}}
%   {\includegraphics[width=\linewidth]{images/SortedwithregionnumberBig}}
% \end{figure}

% \begin{figure}[t]
% \centering
% \includegraphics[width=\linewidth]{images/SortedwithregionnumberBig}
% \caption{Disparity in components selected for SZ and HC subjects. Figure shows the probability density of selected components. Left side has sorted components for SZ patients and right has sorted components for HC subjects. }
% \label{fig:regions}
% \end{figure}

\section{Discussion \& Future work}
We present a novel and relatively shallow deep learning architecture which provides state-of-the-art classification results for schizophrenia. 
Our model learns the functional connectivity between rsfMRI ICA components which can be used as a graph structure and hence result in an individual graph for each subject. Learning connectivity/graph structure in an end to end fashion helps us to not use a fixed method, as such methods does not provide high classification or interpretable results. 

The high classification result we obtained provides evidence that the functional connectivity and sparse components learned by the model are correct and are important, especially for classification.

For future work, we plan to look into the final regions selected by the pooling method and compare those regions across healthy and schizophrenic patients. We also plan to apply our method directly on rs-fMRI to not only reduce but also find disease specific regions.

\acks{Data for Schizophrenia classification was used in this study were acquired from the Function
BIRN Data Repository (http://bdr.birncommunity.org:8080/BDR/),
supported by grants to the Function BIRN (U24-RR021992) Testbed funded by the
National Center for Research Resources at the National Institutes of Health, U.S.A. and from the Collaborative Informatics and Neuroimaging
Suite Data Exchange tool (COINS; \href{http://coins.trendscenter.org}{http://coins.trendscenter.org}). This work was in part supported by NIH grants 1RF1MH12188 and 2R01EB006841.}

\bibliography{jmlr-sample}

\begin{thebibliography}{25}
\providecommand{\natexlab}[1]{#1}
\providecommand{\url}[1]{\texttt{#1}}
\expandafter\ifx\csname urlstyle\endcsname\relax
  \providecommand{\doi}[1]{doi: #1}\else
  \providecommand{\doi}{doi: \begingroup \urlstyle{rm}\Url}\fi

\bibitem[Allen et~al.(2011)Allen, Erhardt, Damaraju, Gruner, Segall, Silva,
  Havlicek, Rachakonda, Fries, Kalyanam, Michael, Caprihan, Turner, Eichele,
  Adelsheim, Bryan, Bustillo, Clark, Feldstein~Ewing, Filbey, Ford, Hutchison,
  Jung, Kiehl, Kodituwakku, Komesu, Mayer, Pearlson, Phillips, Sadek, Stevens,
  Teuscher, Thoma, and Calhoun]{10.3389/fnsys.2011.00002}
Elena Allen, Erik Erhardt, Eswar Damaraju, William Gruner, Judith Segall,
  Rogers Silva, Martin Havlicek, Srinivas Rachakonda, Jill Fries, Ravi
  Kalyanam, Andrew Michael, Arvind Caprihan, Jessica Turner, Tom Eichele,
  Steven Adelsheim, Angela Bryan, Juan Bustillo, Vincent Clark, Sarah
  Feldstein~Ewing, Francesca Filbey, Corey Ford, Kent Hutchison, Rex Jung, Kent
  Kiehl, Piyadasa Kodituwakku, Yuko Komesu, Andrew Mayer, Godfrey Pearlson,
  John Phillips, Joseph Sadek, Michael Stevens, Ursina Teuscher, Robert Thoma,
  and Vince Calhoun.
\newblock A baseline for the multivariate comparison of resting-state networks.
\newblock \emph{Frontiers in Systems Neuroscience}, 5:\penalty0 2, 2011.
\newblock ISSN 1662-5137.
\newblock \doi{10.3389/fnsys.2011.00002}.
\newblock URL
  \url{https://www.frontiersin.org/article/10.3389/fnsys.2011.00002}.

\bibitem[Andreasen and Pierson(2008)]{article3}
Nancy Andreasen and Ronald Pierson.
\newblock The role of the cerebellum in schizophrenia.
\newblock \emph{Biological psychiatry}, 64:\penalty0 81--8, 08 2008.
\newblock \doi{10.1016/j.biopsych.2008.01.003}.

\bibitem[Bianchi et~al.(2021)Bianchi, Grattarola, Livi, and Alippi]{9336270}
Filippo~Maria Bianchi, Daniele Grattarola, Lorenzo Livi, and Cesare Alippi.
\newblock Graph neural networks with convolutional arma filters.
\newblock \emph{IEEE Transactions on Pattern Analysis and Machine
  Intelligence}, pages 1--1, 2021.
\newblock \doi{10.1109/TPAMI.2021.3054830}.

\bibitem[Ebdrup et~al.(2010)Ebdrup, Glenthøj, Rasmussen, Aggernaes, Langkilde,
  Paulson, Lublin, Skimminge, and Baaré]{article2}
Bjørn Ebdrup, Birte Glenthøj, Hans Rasmussen, Bodil Aggernaes, Annika
  Langkilde, Olaf Paulson, Henrik Lublin, Arnold Skimminge, and William Baaré.
\newblock Hippocampal and caudate volume reductions in antipsychotic-naive
  first-episode schizophrenia.
\newblock \emph{Journal of psychiatry \& neuroscience : JPN}, 35:\penalty0
  95--104, 03 2010.
\newblock \doi{10.1503/jpn.090049}.

\bibitem[Fu et~al.(2019)Fu, Caprihan, Chen, Du, Adair, Sui, Rosenberg, and
  Calhoun]{fu2019altered}
Zening Fu, Arvind Caprihan, Jiayu Chen, Yuhui Du, John~C Adair, Jing Sui,
  Gary~A Rosenberg, and Vince~D Calhoun.
\newblock Altered static and dynamic functional network connectivity in
  alzheimer's disease and subcortical ischemic vascular disease: shared and
  specific brain connectivity abnormalities.
\newblock \emph{Human Brain Mapping}, 2019.
\newblock \doi{10.1002/hbm.24591}.

\bibitem[Fu et~al.(2021)Fu, Sui, Turner, Du, Assaf, Pearlson, and
  Calhoun]{https://doi.org/10.1002/hbm.25205}
Zening Fu, Jing Sui, Jessica~A. Turner, Yuhui Du, Michal Assaf, Godfrey~D.
  Pearlson, and Vince~D. Calhoun.
\newblock Dynamic functional network reconfiguration underlying the
  pathophysiology of schizophrenia and autism spectrum disorder.
\newblock \emph{Human Brain Mapping}, 42\penalty0 (1):\penalty0 80--94, 2021.
\newblock \doi{https://doi.org/10.1002/hbm.25205}.
\newblock URL \url{https://onlinelibrary.wiley.com/doi/abs/10.1002/hbm.25205}.

\bibitem[Gao and Ji(2019)]{gao2019graph}
Hongyang Gao and Shuiwang Ji.
\newblock Graph u-nets, 2019.

\bibitem[Jones et~al.(2012)Jones, Vemuri, Murphy, Gunter, Senjem, Machulda,
  Przybelski, Gregg, Kantarci, Knopman, Boeve, Petersen, and Jack]{article}
David Jones, Prashanthi Vemuri, Matthew Murphy, Jeffrey Gunter, Matthew Senjem,
  Mary Machulda, Scott Przybelski, Brian Gregg, Kejal Kantarci, David Knopman,
  Brad Boeve, Ronald Petersen, and Clifford Jack.
\newblock Non-stationarity in the “resting brain’s” modular architecture.
\newblock \emph{PloS one}, 7:\penalty0 e39731, 06 2012.
\newblock \doi{10.1371/journal.pone.0039731}.

\bibitem[Kazi et~al.(2018)Kazi, krishna, Shekarforoush, Kortuem, Albarqouni,
  and Navab]{kazi2018selfattention}
Anees Kazi, S.~Arvind krishna, Shayan Shekarforoush, Karsten Kortuem, Shadi
  Albarqouni, and Nassir Navab.
\newblock Self-attention equipped graph convolutions for disease prediction,
  2018.

\bibitem[Keator et~al.(2016)Keator, van Erp, Turner, Glover, Mueller, Liu,
  Voyvodic, Rasmussen, Calhoun, Lee, et~al.]{keator2016function}
David~B Keator, Theo~GM van Erp, Jessica~A Turner, Gary~H Glover, Bryon~A
  Mueller, Thomas~T Liu, James~T Voyvodic, Jerod Rasmussen, Vince~D Calhoun,
  Hyo~Jong Lee, et~al.
\newblock The function biomedical informatics research network data repository.
\newblock \emph{Neuroimage}, 124:\penalty0 1074--1079, 2016.
\newblock \doi{10.1016/j.neuroimage.2015.09.003}.

\bibitem[Khosla et~al.(2019)Khosla, Jamison, Ngo, Kuceyeski, and
  Sabuncu]{khosla2019machine}
Meenakshi Khosla, Keith Jamison, Gia~H Ngo, Amy Kuceyeski, and Mert~R Sabuncu.
\newblock Machine learning in resting-state {fMRI} analysis.
\newblock \emph{Magnetic resonance imaging}, 2019.

\bibitem[Kipf et~al.(2018)Kipf, Fetaya, Wang, Welling, and
  Zemel]{kipf2018neural}
Thomas Kipf, Ethan Fetaya, Kuan-Chieh Wang, Max Welling, and Richard Zemel.
\newblock Neural relational inference for interacting systems, 2018.

\bibitem[Kipf and Welling(2017)]{kipf2017semisupervised}
Thomas~N. Kipf and Max Welling.
\newblock Semi-supervised classification with graph convolutional networks,
  2017.

\bibitem[Knyazev et~al.(2019)Knyazev, Taylor, and
  Amer]{knyazev2019understanding}
Boris Knyazev, Graham~W. Taylor, and Mohamed~R. Amer.
\newblock Understanding attention and generalization in graph neural networks,
  2019.

\bibitem[Lynall et~al.(2010)Lynall, Bassett, Kerwin, McKenna, Kitzbichler,
  Muller, and Bullmore]{Lynall9477}
Mary-Ellen Lynall, Danielle~S. Bassett, Robert Kerwin, Peter~J. McKenna,
  Manfred Kitzbichler, Ulrich Muller, and Ed~Bullmore.
\newblock Functional connectivity and brain networks in schizophrenia.
\newblock \emph{Journal of Neuroscience}, 30\penalty0 (28):\penalty0
  9477--9487, 2010.
\newblock ISSN 0270-6474.
\newblock \doi{10.1523/JNEUROSCI.0333-10.2010}.
\newblock URL \url{https://www.jneurosci.org/content/30/28/9477}.

\bibitem[Mahmood et~al.(2020)Mahmood, Rahman, Fedorov, Lewis, Fu, Calhoun, and
  Plis]{2020}
Usman Mahmood, Md~Mahfuzur Rahman, Alex Fedorov, Noah Lewis, Zening Fu,
  Vince~D. Calhoun, and Sergey~M. Plis.
\newblock Whole milc: Generalizing learned dynamics across tasks, datasets, and
  populations.
\newblock \emph{Lecture Notes in Computer Science}, page 407–417, 2020.
\newblock ISSN 1611-3349.
\newblock \doi{10.1007/978-3-030-59728-3_40}.
\newblock URL \url{http://dx.doi.org/10.1007/978-3-030-59728-3_40}.

\bibitem[Mahmood et~al.(2021)Mahmood, Fu, Calhoun, and Plis]{2021}
Usman Mahmood, Zening Fu, Vince~D. Calhoun, and Sergey Plis.
\newblock A deep learning model for data-driven discovery of functional
  connectivity.
\newblock \emph{Algorithms}, 14\penalty0 (3):\penalty0 75, Feb 2021.
\newblock ISSN 1999-4893.
\newblock \doi{10.3390/a14030075}.
\newblock URL \url{http://dx.doi.org/10.3390/a14030075}.

\bibitem[Parisot et~al.(2018)Parisot, Ktena, Ferrante, Lee, Guerrero, Glocker,
  and Rueckert]{PARISOT2018117}
Sarah Parisot, Sofia~Ira Ktena, Enzo Ferrante, Matthew Lee, Ricardo Guerrero,
  Ben Glocker, and Daniel Rueckert.
\newblock Disease prediction using graph convolutional networks: Application to
  autism spectrum disorder and alzheimer’s disease.
\newblock \emph{Medical Image Analysis}, 48:\penalty0 117--130, 2018.
\newblock ISSN 1361-8415.
\newblock \doi{https://doi.org/10.1016/j.media.2018.06.001}.
\newblock URL
  \url{https://www.sciencedirect.com/science/article/pii/S1361841518303554}.

\bibitem[Plis et~al.(2014)Plis, Hjelm, Salakhutdinov, Allen, Bockholt, Long,
  Johnson, Paulsen, Turner, and Calhoun]{frontiers2014}
Sergey~M Plis, Devon Hjelm, Ruslan Salakhutdinov, Elena~A Allen, Henry~Jeremy
  Bockholt, Jeffrey~D Long, Hans~J Johnson, Jane Paulsen, Jessica~A Turner, and
  Vince~D Calhoun.
\newblock Deep learning for neuroimaging: a validation study.
\newblock \emph{Frontiers in Neuroscience}, 8\penalty0 (229), 2014.
\newblock ISSN 1662-453X.

\bibitem[Schuster and Paliwal(1997)]{650093}
M.~Schuster and K.K. Paliwal.
\newblock Bidirectional recurrent neural networks.
\newblock \emph{IEEE Transactions on Signal Processing}, 45\penalty0
  (11):\penalty0 2673--2681, 1997.
\newblock \doi{10.1109/78.650093}.

\bibitem[Shang et~al.(2021)Shang, Chen, and Bi]{shang2021discrete}
Chao Shang, Jie Chen, and Jinbo Bi.
\newblock Discrete graph structure learning for forecasting multiple time
  series.
\newblock In \emph{International Conference on Learning Representations}, 2021.
\newblock URL \url{https://openreview.net/forum?id=WEHSlH5mOk}.

\bibitem[Vaswani et~al.(2017)Vaswani, Shazeer, Parmar, Uszkoreit, Jones, Gomez,
  Kaiser, and Polosukhin]{10.5555/3295222.3295349}
Ashish Vaswani, Noam Shazeer, Niki Parmar, Jakob Uszkoreit, Llion Jones,
  Aidan~N. Gomez, undefinedukasz Kaiser, and Illia Polosukhin.
\newblock Attention is all you need.
\newblock In \emph{Proceedings of the 31st International Conference on Neural
  Information Processing Systems}, NIPS'17, page 6000–6010, Red Hook, NY,
  USA, 2017. Curran Associates Inc.
\newblock ISBN 9781510860964.

\bibitem[Ying et~al.(2018)Ying, You, Morris, Ren, Hamilton, and
  Leskovec]{NEURIPS2018_e77dbaf6}
Zhitao Ying, Jiaxuan You, Christopher Morris, Xiang Ren, Will Hamilton, and
  Jure Leskovec.
\newblock Hierarchical graph representation learning with differentiable
  pooling.
\newblock In S.~Bengio, H.~Wallach, H.~Larochelle, K.~Grauman, N.~Cesa-Bianchi,
  and R.~Garnett, editors, \emph{Advances in Neural Information Processing
  Systems}, volume~31. Curran Associates, Inc., 2018.
\newblock URL
  \url{https://proceedings.neurips.cc/paper/2018/file/e77dbaf6759253c7c6d0efc5690369c7-Paper.pdf}.

\bibitem[Zhang and Chen(2018)]{zhang2018link}
Muhan Zhang and Yixin Chen.
\newblock Link prediction based on graph neural networks, 2018.

\bibitem[Zhang et~al.(2018)Zhang, Cui, Neumann, and Chen]{zhang2018end}
Muhan Zhang, Zhicheng Cui, Marion Neumann, and Yixin Chen.
\newblock An end-to-end deep learning architecture for graph classification.
\newblock In \emph{AAAI}, pages 4438--4445, 2018.

\end{thebibliography}

% \appendix

% \section{First Appendix}\label{apd:first}

% This is the first appendix.

% \section{Second Appendix}\label{apd:second}

% This is the second appendix.

\end{document}